\documentclass[11pt,a4paper,dvipsnames]{article}
\usepackage[hyperref]{acl2018}
\usepackage{times}
\usepackage{latexsym}
\usepackage{booktabs}
\usepackage{amsmath}
\usepackage{amssymb}
\usepackage{url}

\usepackage{graphicx}
\usepackage{tikz}

\newcommand{\nl}[1]{\textit{{\small #1}}}
\makeatletter
\newcommand*\bigcdot{\mathpalette\bigcdot@{.5}}
\newcommand*\bigcdot@[2]{\mathbin{\vcenter{\hbox{\scalebox{#2}{$\m@th#1\bullet$}}}}}
\makeatother

\usetikzlibrary{shapes.geometric} 
\usetikzlibrary{arrows, backgrounds, calc, hobby, positioning}

\ExplSyntaxOn
\cs_if_exist:NF \prg_stepwise_function:nnnN { \cs_gset_eq:NN \prg_stepwise_function:nnnN \int_step_function:nnnN }
\cs_if_exist:NF \prg_stepwise_inline:nnnn { \cs_gset_eq:NN \prg_stepwise_inline:nnnn \int_step_inline:nnnn }
\ExplSyntaxOff

\tikzset{vertex style/.style={
    draw=#1,
    thick,
    fill=#1!70,
    text=white,
    ellipse,
    minimum width=2cm,
    minimum height=0.75cm,
    font=\small,
    outer sep=3pt,
  },
  text style/.style={
    sloped,
    text=black,
    font=\footnotesize,
    above
  }
}

\aclfinalcopy 
\def\aclpaperid{***} 

\title{Deep Graph Convolutional Encoders for \\Structured Data to Text Generation}

\author{
 Diego Marcheggiani$^{1,2}$ \hspace{1cm} Laura Perez-Beltrachini$^{1}$\\
 $^1$ILCC, School of Informatics, University of Edinburgh \\
 $^2$ILLC,  University of Amsterdam  \\
    {\tt marcheggiani@uva.nl} \hspace{0.3cm} {\tt {lperez}@inf.ed.ac.uk} 
 }
\date{}

\begin{document}
\maketitle
\begin{abstract}
Most previous work on neural text generation from graph-structured 
data relies on standard sequence-to-sequence methods. These approaches
linearise the input graph to be fed to a recurrent neural network.
In this paper, we propose an alternative encoder based on graph 
convolutional networks that directly exploits the input structure.
We report results on two graph-to-sequence datasets that empirically 
show the benefits of explicitly encoding the input graph structure.\footnote{Code and data available at \url{github.com/diegma/graph-2-text}.}

\end{abstract}

\section{Introduction}

Data-to-text generators produce a target natural language text 
from a source data representation. Recent neural generation approaches 
\cite{mei2015talk,lebret-grangier-auli:2016:EMNLP2016,wiseman-shieber-rush:2017:EMNLP2017,gardent-EtAl:2017:INLG2017,ferreira2017linguistic,konstas2017neural} 
build on encoder-decoder architectures proposed for machine translation
\cite{sutskever2014sequence,bahdanau2015neural}.

\begin{figure*}[ht]
\begin{center}
\begin{minipage}[t]{0.55\linewidth}
\vspace{0pt}
\begin{footnotesize}
  \begin{center}
\begin{tikzpicture}[node distance=2.75cm,>=stealth',scale=0.6,  every node/.style={scale=0.6}]
\node[vertex style=Turquoise] (Rk) {Above the Veil};

\node[vertex style=Turquoise, below right of=Rk,xshift=2em,yshift=-1em] (JA) {Aenir}
 edge [<-,cyan!60!blue] node[text style]{precededBy} (Rk); 
 
\node[vertex style=Turquoise, below right of=JA,xshift=-2em,yshift=-2em] (CA) {Castle}
 edge [<-,cyan!60!blue] node[text style]{precededBy} (JA);  

 \node[vertex style=BurntOrange, right=1.5cm of Rk,yshift=-4ex] (RN) {Australians}
 edge [<-,cyan!60!blue] node[text style]{country} (Rk); 

\node[vertex style=Turquoise, below of=Rk,xshift=-2em] (Skf) {Into Battle}
 edge [<-,cyan!60!blue] node[text style]{followedBy} (Rk);

\node[vertex style=Turquoise, below right of=Skf,xshift=-2em,yshift=-2em] (Cf) {The Violet Keystone}
 edge [<-,cyan!60!blue] node[text style]{followedBy} (Skf);

\end{tikzpicture}
\vspace{-0.5em}
  \end{center}
 \end{footnotesize}
  (a) \nl{Above the Veil is an Australian novel and the sequel to Aenir and Castle . It was followed by Into the Battle and The Violet Keystone .}
\end{minipage}
\vspace{0pt}
\begin{minipage}[t]{0.43\linewidth}
\vspace{0pt}
\centering
    \includegraphics[scale=0.42]{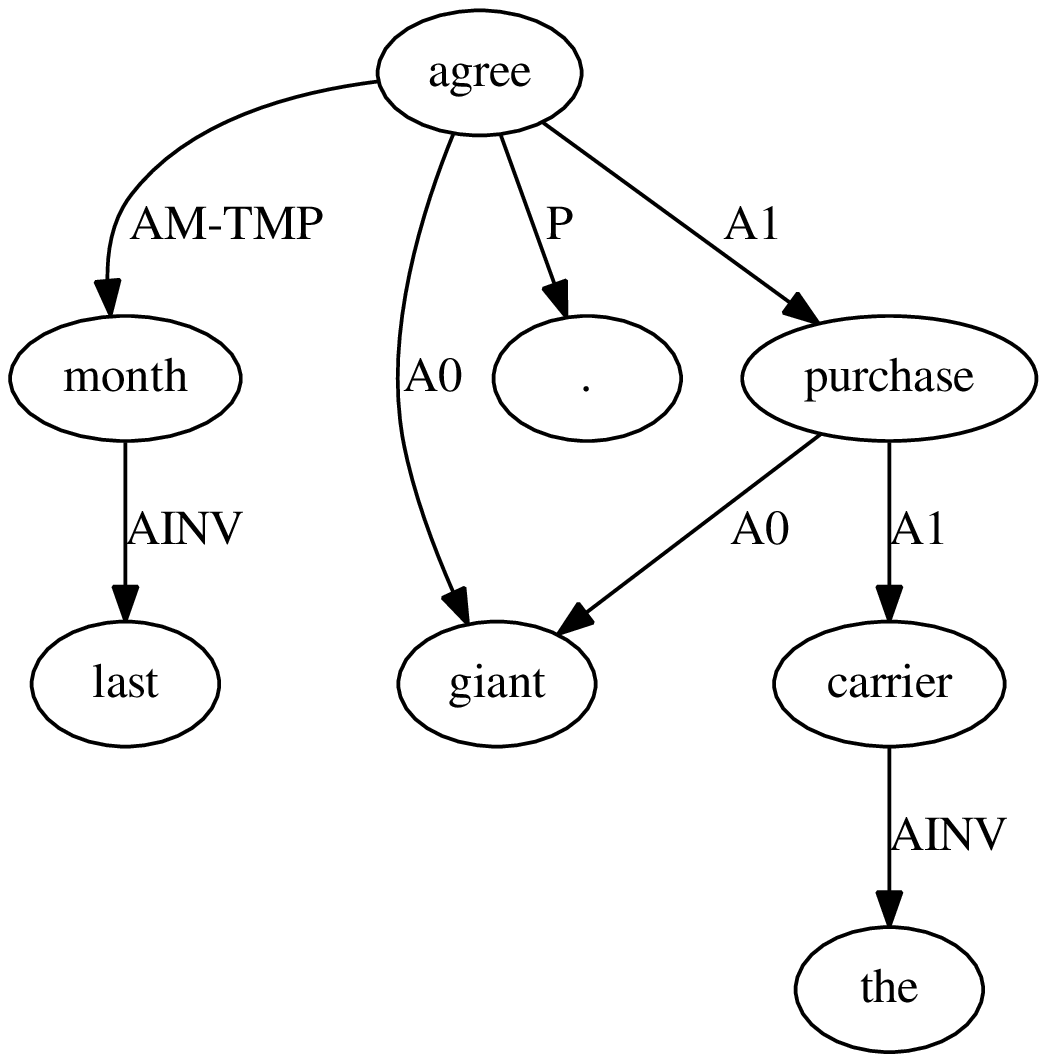}
\\
\vspace{-0.5em}
(b) \nl{Giant agreed last month to purchase the carrier . }
\end{minipage}
 
\end{center}
\vspace{-1em}
\caption{Source RDF graph - target description (a). Source dependency graph - target sentence (b).}\label{fig:rdfgraph_srgraph}
\end{figure*}

The source data, differently from the machine translation task, 
is a structured representation of the content to be conveyed.
Generally, it describes attributes and events about entities and relations among them.
In this work we focus on two generation scenarios where the source
data is graph structured.
One is the generation of multi-sentence descriptions
of Knowledge Base (KB) entities from RDF graphs
\cite{perezbeltrachini-sayed-gardent:2016:COLING,gardent-EtAl:2017:Long,gardent-EtAl:2017:INLG2017},
namely the WebNLG task.\footnote{Resource Description Framework \url{https://www.w3.org/RDF/}}
The number of KB relations modelled in this scenario is potentially large and generation involves 
solving various subtasks (e.g. lexicalisation and 
aggregation). Figure (\ref{fig:rdfgraph_srgraph}a) shows and example 
of source RDF graph and target natural language description.
The other is the linguistic realisation of the meaning expressed 
by a source dependency graph \cite{belz2011first}, namely the SR11Deep generation task. 
In this task, the semantic relations are linguistically motivated and their number 
is smaller.
Figure (\ref{fig:rdfgraph_srgraph}b) illustrates a source dependency graph 
and the corresponding target text.

Most previous work casts the graph structured data to text generation 
task as a sequence-to-sequence problem \cite{gardent-EtAl:2017:INLG2017,ferreira2017linguistic,konstas2017neural}.
They rely on recurrent data encoders with memory and gating
mechanisms (LSTM; \cite{lstm1997}). 
Models based on these sequential encoders have shown 
good results although they do not directly exploit
the input structure but rather rely on a separate linearisation step. 
In this work, we compare with a model that explicitly encodes 
structure and is trained end-to-end.
Concretely, we use a Graph Convolutional Network 
(GCN; \cite{kipf2016semi,marcheggiani2017encoding}) 
as our encoder.

GCNs are a flexible architecture that allows explicit encoding 
of graph data into neural networks. Given their simplicity and 
expressiveness they have been used to encode dependency syntax 
and predicate-argument structures in neural machine translation 
\cite{bastings2017graph,marcheggiani2018exploiting}.
In contrast to previous work, we do not exploit the sequential 
information of the input (i.e., with an LSTM), but we solely rely 
on a GCN for encoding the source graph structure.\footnote{Concurrently with this work, \newcite{P18-1026} also encoded input structures without relying on sequential encoders.}

The main contribution of this work is showing that explicitly encoding structured data with GCNs is more effective than encoding a linearized version of the structure with LSTMs.
We evaluate the GCN-based generator on two 
graph-to-sequence tasks, with different level of source content specification.
In both cases, the results we obtain show that GCNs encoders outperforms standard 
LSTM encoders.

\section{Graph Convolutional-based Generator}
\label{sec:d2tmodel}

Formally, we address the task of text generation from 
graph-structured data considering as input a directed labeled 
graph~$X=(\mathcal{V}, \mathcal{E})$ where $\mathcal{V}$ is a set 
of nodes and $\mathcal{E}$ is a set of edges between nodes 
in $\mathcal{V}$. The specific semantics of $X$ depends on the task at hand. 
The output $Y$ is a natural language text verbalising the content expressed by $X$.
Our generation model follows the standard attention-based encoder-decoder 
architecture \cite{bahdanau2015neural,luong2015effective} and predicts $Y$
conditioned on $X$ as $ P(Y|X) = \prod_{t=1}^{|Y|} P(y_t|y_{1:t-1}, X)$.

\paragraph{Graph Convolutional Encoder}
In order to explicitly encode structural information we adopt graph convolutional networks (GCNs).
GCNs are a variant of graph neural networks \cite{scarselli2009graph} that has been recently proposed by \newcite{kipf2016semi}.
The goal of GCNs is to calculate the representation of each node in a graph considering the graph structure.
In this paper we adopt the parametrization proposed by \newcite{marcheggiani2017encoding} where edge labels and directions are explicitly modeled.
Formally, given a directed graph $X = (\mathcal{V}, \mathcal{E})$, where $\mathcal{V}$ is a set of nodes, and $\mathcal{E}$ is a set of edges. 
We represent each node $v \in \mathcal{V}$ with a feature vector $\mathbf{x}_v \in \mathbb{R}^d$.
The GCN calculates the representation of each node $\mathbf{h}'_v$ in a graph using the following update rule:
\vspace{-0.5em}
\begin{align*}\label{eq:gcn}
\mathbf{h}'_v\!\!=&\rho\Big(\!\!\!\sum_{u \in \mathcal{N}(v)} \!\!\! g_{u, v} \big(W_{dir(u, v)} \, \mathbf{h}_u + \mathbf{b}_{lab(u, v)}\big)\Big),
\end{align*}
\noindent
where $\mathcal{N}(v)$ is the set of neighbours of $v$, $W_{dir(u, v)} \in \mathbb{R}^{d \times d}$ is a direction-specific parameter matrix. 
As \newcite{marcheggiani2017encoding,bastings2017graph} we assume there are three possible directions ($dir(u, v) \in \{in, out, loop\}$): self-loop edges ensure that the initial representation of node $\mathbf{h}_v$ affects the new representation $\mathbf{h}'_v$. 
The vector $\mathbf{b}_{lab(u, v)} \in \mathbb{R}^d$ is an embedding of the label of the edge $(u,v)$ 
. $\rho$ is a non-linearity (ReLU).
$g_{u, v}$ are learned scalar gates which weight the importance of each edge.
Although the main aim of gates is to down weight erroneous edges in predicted graphs, they also add flexibility when several GCN layers are stacked.
As with standard convolutional neural networks (CNNs, \cite{lecun-01a}), GCN layers can be stacked to consider non-immediate neighbours.\footnote{We discovered during preliminary experiments that without scalar gates the model ends up in poor local minima, especially when several GCN layers are used.} 

\paragraph{Skip Connections} Between GCN layers we add skip connections.
Skip connections let the gradient flows more efficiently through stacked hidden layers thus making possible the creation of deeper GCN encoders.
We use two kinds of skip connections: residual connections \cite{he2016deep} and dense connections \cite{DBLP:conf/cvpr/HuangLMW17}.
Residual connections consist in summing input and output representations of a GCN layer 
$\mathbf{h}^r_v \!=\! \mathbf{h}'_v \! + \! \mathbf{h}_v$.
Whilst, dense connections consist in the concatenation of the input and output representations $\mathbf{h}^d_v = [\mathbf{h}'_v;\mathbf{h}_v]$.
In this way, each GCN layer is directly fed with the output of every layer before itself.

\vspace*{-0.5ex}
\paragraph{Decoder}
The decoder uses an LSTM and a soft attention mechanism \cite{luong2015effective} over the representation induced by the GCN encoder to generate one word~$y$ at the time.
The prediction of word $y_{t+1}$ is conditioned on the previously predicted words $y_{1:t}$ encoded in the vector $\mathbf{w}_t$ and a context vector $\mathbf{c}_t$
dynamically created attending to the graph representation induced by the GCN encoder as
$ P(y_{t+1}|y_{1:t}, X) = softmax(g(\mathbf{w}_t, \mathbf{c}_t))$,
where $g(\cdot)$ is a neural network with one hidden layer.
The model is trained to optimize negative log likelihood:
${\cal L}_{NLL} =  - \sum_{t=1}^{|Y|} log \, P(y_t|y_{1:t-1}, X)$

\section{Generation Tasks}
\label{sec:d2ttasks}
In this section, we describe the instantiation of the input graph $X$
for the generation tasks we address.

\subsection{WebNLG Task}
The WebNLG task \cite{gardent-EtAl:2017:Long,gardent-EtAl:2017:INLG2017} 
aims at the generation of entity descriptions from a set of RDF triples 
related to an entity of a given category \cite{perezbeltrachini-sayed-gardent:2016:COLING}. 
RDF triples are of the form ({\sffamily subject relation object}), e.g.,
({\sffamily Aenir precededBy Castle}), and form a graph in which edges are 
labelled with relations and vertices with subject and object entities.
For instance, Figure (\ref{fig:rdfgraph_srgraph}a) shows a set of RDF 
triples related to the book \emph{Above the Veil} and its verbalisation.
The generation task involves several micro-planning decisions such as lexicalisation ({\sffamily followedBy}
is verbalised as \nl{sequel to}), aggregation (\nl{sequel to Aenir and Castle}),
referring expressions (subject of the second sentence verbalised as pronoun)
and segmentation (content organised in two sentences).

\paragraph{Reification}
We formulate this task as the generation of a target description $Y$ 
from a source graph $X=(\mathcal{V}, \mathcal{E})$ where $X$ 
is build from a set of RDF triples as follows. 
We reify the relations \cite{baader2003description} from the RDF set of 
triples. That is, we see the relation as a concept in the KB and
introduce a new relation node for each relation of each RDF triple.
The new relation node is connected to the subject and object entities 
by two new binary relations A0 and A1 respectively.
For instance, ({\sffamily precededBy A0 Aenir}) and ({\sffamily precededBy A1 Castle}).
Thus, $\mathcal{E}$ is the set of entities including reified relations
and  $\mathcal{V}$ a set of labelled edges with labels $\{A0, A1\}$.
The reification of relations is useful in two ways. The encoder is able to 
produce a hidden state for each relation in the input; and it permits to model
an arbitrary number of KB relations efficiently.

\subsection{SR11Deep Task}
The surface realisation shared task \cite{belz2011first} 
proposed two generation tasks, namely shallow and deep realisation.
Here we focus on the deep task where the input is a semantic dependency graph that
represents a target sentence using predicate-argument structures
(NomBank; \cite{meyers2004nombank}, PropBank; \cite{palmer2005proposition}).
This task covers a more complex semantic representation of
language meaning; on the other hand, the representation is closer
to surface form. 
Nodes in the graph are lemmas of the target sentence. Only complementizers \nl{that}, commas, and \nl{to}
infinitive nodes are removed. Edges are labelled with NomBank
and PropBank labels.\footnote{There are also some cases where syntactic
 labels appear in the graphs, this is due to the creation process (see \cite{belz2011first})
 and done to connect graphs when there were disconnected parts.}
Each node is also associated with morphological (e.g. num=sg) and
punctuation features (e.g. bracket=r). 

The source graph $X=(\mathcal{V}, \mathcal{E})$ is a semantic 
dependency graph. We extend this representation to model morphological
information, i.e. each node in $\mathcal{V}$ is of the form
(lemma, features). For this task we modify the encoder, Section~\ref{sec:d2tmodel},
to represent each input node as $\mathbf{h}_v= [\mathbf{h}_l;\mathbf{h}_f] $,
where each input node is the concatenation of the lemma and the sum of feature
vectors.

\section{Experiments}
\label{sec:evalsetup}

We tested our models on the WebNLG and SR11Deep datasets.
The WebNLG dataset contains 18102 training and 871 development 
data-text pairs. 
The test dataset is split in two sets, test \textit{Seen} (971 pairs)
and a test set with new unseen categories for KB entities.
As here we are interested only in the modelling aspects of the 
structured input data we focus on our evaluation only on the 
test partition with seen categories.
The dataset covers 373 distinct relations from DBPedia. 
The SR11Deep dataset contains 39279, 1034 and 2398 examples in 
the training, development and test partitions, respectively.
It covers 117 distinct dependency relations.\footnote{
In both datasets we exclude pairs with $>$50 target words.}

\paragraph{Sequential Encoders}
For both WebNLG and SR11Deep tasks we used a standard sequence-to-sequence model \cite{bahdanau2015neural,luong2015effective} with an LSTM encoder as baseline.
Both take as input a linearised version of the source graph.
For the WebNLG baseline, we use the linearisation scripts provided by 
\cite{gardent-EtAl:2017:INLG2017}. 
For the SR11Deep baseline we follow a similar linearisation procedure
as proposed for AMR graphs \cite{konstas2017neural}.
We built a linearisation based on a depth first traversal of the input graph. 
Siblings are traversed in random order (they are anyway shuffled in the given 
dataset). 
We repeat a child node when a node is revisited by a cycle
or has more than one parent.
The baseline model for the WebNLG task uses one layer bidirectional LSTM encoder and one layer LSTM decoder with embeddings and hidden units set to 256 dimensions . 
For the SR11Deep task we used the same architecture with 500-dimensional hidden states and embeddings.
All hyperparameters tuned on the development set.

\paragraph{GCN Encoders}
The GCN models consist of a GCN encoder and LSTM decoder.
For the WebNLG task, all encoder and decoder embeddings and hidden units
use 256 dimensions. 
We obtained the best results with an encoder with four GCN layers with residual connections.
For the SR11Deep task, we set the encoder and decoder to use 500-dimensional 
embeddings and hidden units of size 500. In this task, we obtained 
the best development performance by stacking seven GCN layers with dense connections.

We use delexicalisation for the WebNLG dataset and apply 
the procedure provided for the baseline in \cite{gardent-EtAl:2017:INLG2017}. 
For the SR11Deep dataset, we performed entity anonymisation. 
First, we compacted nodes in the tree corresponding to a single 
named entity (see \cite{belz2011first} for details). 
Next, we used a name entity recogniser (Stanford CoreNLP; 
\cite{manning-EtAl:2014:P14-5}) to tag entities in the input 
with type information (e.g. person, location, date). Two 
entities of the same type in a given input will be given
a numerical suffix, e.g. PER\_0 and PER\_1.

\paragraph{A GCN-based Generator}
For the WebNLG task, we extended the GCN-based model to use pre-trained word \textbf{E}mbeddings (GloVe \cite{pennington2014glove})
and \textbf{C}opy mechanism \cite{P17-1099}, we name this variant GCN$_{EC}$. 
To this end, we did not use delexicalisation but rather
represent multi-word subject (object) entities with each word
as a separate node connected with special Named Entity (NE) labelled edges.
For instance, the book entity \emph{Into Battle} is represented as {\sffamily(Into NE Battle)}.
Encoder (decoder) embeddings and hidden dimensions were set to 300.
The model stacks six GCN layers and uses a single layer LSTM decoder.

\paragraph{Evaluation metrics}
As previous works in these tasks,
we evaluated our models using BLEU \cite{papineni2002bleu}, 
METEOR \cite{denkowski2014meteor} and TER \cite{snover2006study}
automatic metrics.
During preliminary experiments we noticed considerable variance 
from different model initialisations; we thus run 3 experiments 
for each model and report average and standard deviation for each metric.

\section{Results}
\label{sec:results}

\begin{table}[t]
\centering
\begin{footnotesize}
\begin{tabular}{@{\extracolsep{\fill}}l@{\hspace{6pt}}l@{\hspace{6pt}}l@{\hspace{6pt}}l@{\hspace{6pt}}l}
  \toprule
   & Encoder & BLEU  & METEOR & TER  \\
   \midrule
   &  LSTM   & .526$\pm$.010 & .38$\pm$.00  & .43$\pm$.01 \\
   &  GCN    & .535$\pm$.004 & .39$\pm$.00 & .44$\pm$.02  \\
   \midrule 
   & \textsc{ADAPT}      & .606 & .44 & .37 \\
    & GCN$_{EC}$ 	 & .559$\pm$.017 & .39$\pm$.01 & 0.41$\pm$.01 \\
   & \textsc{Melbourne}  & .545 & .41 & .40 \\
   & \textsc{PKUWriter}  & .512 & .37 & .45 \\
\bottomrule
\end{tabular}
\end{footnotesize}
\vspace{-0.5em}
\caption{\label{tab:test-results-webnlg} Test results WebNLG task.}
\end{table}

\begin{table}[t]
\centering
\begin{footnotesize}
\begin{tabular}{@{\extracolsep{\fill}}l@{\hspace{6pt}}l@{\hspace{6pt}}l@{\hspace{6pt}}l@{\hspace{6pt}}l}
  \toprule
   & Encoder & BLEU  & METEOR & TER  \\
   \midrule
   & LSTM & .377$\pm$.007 & .65$\pm$.00 & .44$\pm$.01 \\
   & GCN  & .647$\pm$.005 & .77$\pm$.00& .24$\pm$.01 \\
   & GCN+feat  & .666$\pm$.027 & .76$\pm$.01 & .25$\pm$.01  \\
\bottomrule
\end{tabular}
\end{footnotesize}
\vspace{-0.5em}
\caption{\label{tab:test-results-sr11} Test results SR11Deep task.}
\end{table}

\paragraph{WebNLG task}

In Table \ref{tab:test-results-webnlg} we report results on the WebNLG test data.
In this setting, the model with GCN encoder outperforms a strong baseline that employs the LSTM encoder, with $.009$ BLEU points. 
The GCN model is also more stable than the baseline with a standard deviation of $.004$ vs $.010$.
We also compared the GCN$_{EC}$ model with the neural models submitted to the WebNLG shared task.
The GCN$_{EC}$ model outperforms \textsc{PKUWriter} that uses an ensemble 
of 7 models and a further reinforcement learning step by $.047$ BLEU points; and \textsc{Melbourne} by $.014$ BLEU points.
GCN$_{EC}$ is behind \textsc{ADAPT} which relies on sub-word encoding.

\begin{table*}[t]
\begin{scriptsize}
 \begin{tabular}{|@{~}p{0.8cm}p{14.5cm}@{~}|}
  \hline
  WebNLG & (William Anders dateOfRetirement 1969 - 09 - 01) (Apollo 8 commander Frank Borman) (William Anders was a crew member of Apollo 8)  (Apollo 8 backup pilot Buzz Aldrin)\\
  LSTM & William Anders was a crew member of the OPERATOR operated Apollo 8 and retired on September 1st 1969 .\\
  GCN & William Anders was a crew member of OPERATOR ' s Apollo 8 alongside backup pilot Buzz Aldrin and backup pilot Buzz Aldrin .\\
  GCN$_{EC}$ & william anders , who retired on the 1st of september 1969 , was a crew member on apollo 8 along with commander frank borman and backup pilot buzz aldrin .\\

  \hline
  SR11Deep & (SROOT SROOT will) (will P .) (will SBJ temperature) (temperature A1 economy) (economy AINV the) (economy SUFFIX 's) (will VC be) (be VC take) (take A1 temperature) (take A2 from) (from A1 point) (point A1 vantage) (point AINV several) (take AM-ADV with) (with A1 reading) (reading A1 on) (on A1 trade) (trade COORD output) (output COORD housing) (housing COORD and) (and CONJ inflation) (take AM-MOD will) (take AM-TMP week) (week AINV this)\\
  Gold & The economy 's temperature will be taken from several vantage points this week , with readings on trade , output , housing and inflation . \\
  Baseline & the economy 's accords will be taken from several phases this week , housing and inflation readings on trade , housing and inflation .\\
  GCN & the economy 's temperatures will be taken from several vantage points this week , with reading on trades output , housing and inflation .\\
\hline
 \end{tabular} 
\end{scriptsize}
\vspace*{-1.5ex}
\caption{\label{tab:sys-output} Examples of system output.}
\end{table*}

\begin{table}[t]
\centering
\begin{footnotesize}
\begin{tabular}{@{\extracolsep{\fill}}l@{\hspace{3pt}}l@{\hspace{4pt}}c@{\hspace{4pt}}c@{\hspace{4pt}}c@{\hspace{5pt}}r@{\hspace{5pt}}c@{\hspace{5pt}}c}
  \toprule
   & & \multicolumn{3}{c}{BLEU}  & \multicolumn{3}{c}{SIZE} \\
   & Model & none  & res & den  & none  & res & den\\
   \midrule
      & LSTM      & .543$\pm$.003 & - & - & 4.3 & - & - \\
      \midrule
      & GCN & \\
      & $\;\;\;$1L   & .537$\pm$.006 & - & -  & 4.3 & - & -   \\
      & $\;\;\;$2L & .545$\pm$.016 & .553$\pm$.005 & .552$\pm$.013  & 4.5 & 4.5 & 4.7 \\
      & $\;\;\;$3L & .548$\pm$.012 & .560$\pm$.013 & .557$\pm$.001  & 4.7 & 4.7 & 5.2 \\
      & $\;\;\;$4L & .537$\pm$.005 & .569$\pm$.003 & .558$\pm$.005  & 4.9 & 4.9 & 6.0 \\
      & $\;\;\;$5L & .516$\pm$.022 & .561$\pm$.016 & .559$\pm$.003  & 5.1 & 5.1 & 7.0 \\
      & $\;\;\;$6L & .508$\pm$.022 & .561$\pm$.007 & .558$\pm$.018  & 5.3 & 5.3 & 8.2 \\
      & $\;\;\;$7L & .492$\pm$.024 & .546$\pm$.023 & .564$\pm$.012  & 5.5 & 5.5 & 9.6 \\

\bottomrule
\end{tabular}
\end{footnotesize}
\vspace{-1.5ex}
\caption{\label{tab:dev-results-webnlg} GCN ablation study  
(layers (L) and skip-connections: none, residual(res) and dense(den)). 
Average and standard deviation of BLEU scores over three runs on the WebNLG dev. set.
Number of parameters (millions) including embeddings. 
}
\end{table}

\paragraph{SR11Deep task}
In this more challenging task, the GCN encoder is able to better capture the structure of the input graph than the LSTM encoder, resulting in $.647$ BLEU for the GCN vs. $.377$ BLEU of the LSTM encoder as reported in Table~\ref{tab:test-results-sr11}.
When we add linguistic features to the GCN encoding we get $.666$ BLEU points.
We also compare the neural models with upper bound results on the same dataset by
the pipeline model of \newcite{DBLP:conf/enlg/BohnetMFW11} 
(STUMBA-D) and transition-based joint model of \newcite{DBLP:conf/eacl/ZhangSP17} (TBDIL).
The STUMBA-D and TBDIL model obtains respectively $.794$ and $.805$ BLUE, outperforming the GCN-based model.
It is worth noting that these models rely on separate modules for 
syntax prediction, tree linearisation and morphology generation.
In a multi-lingual setting \cite{mille2017shared},
our model will not need to re-train some modules for different languages, 
but rather it can exploit them for multi-task training.
Moreover, our model could also exploit other supervision signals at training time, 
such as gold POS tags and gold syntactic trees as used in \newcite{DBLP:conf/enlg/BohnetMFW11}.

\subsection{Qualitative Analysis of Generated Text}
We manually inspected the outputs of the LSTM and GCN models.
Table~\ref{tab:sys-output} shows examples of source graphs and generated texts 
(we included more examples in Section \ref{sec:supMaterial}).
Both models suffer from repeated and missing source content (i.e. source 
units are not verbalised in the output text (under-generation)).
However, these phenomena are less evident with GCN-based models. 
We also observed that the LSTM output sometimes presents hallucination 
(over-generation) cases.
Our intuition is that the strong relational inductive bias of GCNs \cite{battaglia2018relational} helps the GCN encoder to produce a more informative representation of the input; while the LSTM-based encoder has to learn to produce useful representations by going through multiple different sequences over the source data.

\subsection{Ablation Study}
In Table \ref{tab:dev-results-webnlg} (BLEU) we report an ablation study on the impact of the number of layers and the type of skip connections on the WebNLG dataset.
The first thing we notice is the importance of skip connections between GCN layers.
Residual and dense connections lead to similar results.
Dense connections (Table \ref{tab:dev-results-webnlg} (SIZE)) produce models bigger, but slightly less accurate, than residual connections.
The best GCN model has slightly more parameters than the baseline model (4.9M vs.4.3M).

\vspace{-0.5em}
\section{Conclusion}
\label{sec:conslusion}
\vspace{-0.5em}
We compared LSTM sequential encoders with a structured 
data encoder based on GCNs on the task of structured data 
to text generation.
On two different tasks, WebNLG and SR11Deep, we
show that explicitly encoding structural information 
with GCNs is beneficial with respect to sequential encoding. 
In future work, we plan to apply the approach to other
input graph representations like Abstract Meaning Representations (AMR; \cite{banarescu2013abstract})
and scoped semantic representations \cite{Noord2018LREC}.

 \vspace{-1ex}
 \section*{Acknowledgments}
 \vspace{-1ex}
We want to thank Ivan Titov and Mirella Lapata for their help and suggestions.
We also gratefully acknowledge the financial support
of the European Research Council (award number 681760)
and the Dutch National Science Foundation (NWO VIDI 639.022.518). 
We thank NVIDIA for donating the GPU used for this research.

\bibliography{gcng2s_inlg2018}

\begin{thebibliography}{40}
\expandafter\ifx\csname natexlab\endcsname\relax\def\natexlab#1{#1}\fi

\bibitem[{Baader(2003)}]{baader2003description}
Franz Baader. 2003.
\newblock \emph{The description logic handbook: Theory, implementation and
  applications}.
\newblock Cambridge university press.

\bibitem[{Bahdanau et~al.(2015)Bahdanau, Cho, and Bengio}]{bahdanau2015neural}
Dzmitry Bahdanau, Kyunghyun Cho, and Yoshua Bengio. 2015.
\newblock \href {http://arxiv.org/abs/1409.0473} {{Neural Machine Translation
  by Jointly Learning to Align and Translate}}.
\newblock In \emph{{Proceedings of the International Conference on Learning
  Representations, ICLR}}.

\bibitem[{Banarescu et~al.(2013)Banarescu, Bonial, Cai, Georgescu, Griffitt,
  Hermjakob, Knight, Koehn, Palmer, and Schneider}]{banarescu2013abstract}
Laura Banarescu, Claire Bonial, Shu Cai, Madalina Georgescu, Kira Griffitt, Ulf
  Hermjakob, Kevin Knight, Philipp Koehn, Martha Palmer, and Nathan Schneider.
  2013.
\newblock Abstract meaning representation for sembanking.
\newblock In \emph{Proceedings of the 7th Linguistic Annotation Workshop and
  Interoperability with Discourse}, pages 178--186.

\bibitem[{Bastings et~al.(2017)Bastings, Titov, Aziz, Marcheggiani, and
  Simaan}]{bastings2017graph}
Joost Bastings, Ivan Titov, Wilker Aziz, Diego Marcheggiani, and Khalil Simaan.
  2017.
\newblock \href {https://www.aclweb.org/anthology/D17-1209} {Graph
  convolutional encoders for syntax-aware neural machine translation}.
\newblock In \emph{Proceedings of the 2017 Conference on Empirical Methods in
  Natural Language Processing, {EMNLP}}, pages 1957--1967.

\bibitem[{Battaglia et~al.(2018)Battaglia, Hamrick, Bapst, Sanchez{-}Gonzalez,
  Zambaldi, Malinowski, Tacchetti, Raposo, Santoro, Faulkner,
  G{\"{u}}l{\c{c}}ehre, Song, Ballard, Gilmer, Dahl, Vaswani, Allen, Nash,
  Langston, Dyer, Heess, Wierstra, Kohli, Botvinick, Vinyals, Li, and
  Pascanu}]{battaglia2018relational}
Peter~W. Battaglia, Jessica~B. Hamrick, Victor Bapst, Alvaro
  Sanchez{-}Gonzalez, Vin{\'{\i}}cius~Flores Zambaldi, Mateusz Malinowski,
  Andrea Tacchetti, David Raposo, Adam Santoro, Ryan Faulkner, {\c{C}}aglar
  G{\"{u}}l{\c{c}}ehre, Francis Song, Andrew~J. Ballard, Justin Gilmer,
  George~E. Dahl, Ashish Vaswani, Kelsey Allen, Charles Nash, Victoria
  Langston, Chris Dyer, Nicolas Heess, Daan Wierstra, Pushmeet Kohli, Matthew
  Botvinick, Oriol Vinyals, Yujia Li, and Razvan Pascanu. 2018.
\newblock \href {http://arxiv.org/abs/1806.01261} {Relational inductive biases,
  deep learning, and graph networks}.
\newblock \emph{CoRR}, abs/1806.01261.

\bibitem[{Beck et~al.(2018)Beck, Haffari, and Cohn}]{P18-1026}
Daniel Beck, Gholamreza Haffari, and Trevor Cohn. 2018.
\newblock \href {http://aclweb.org/anthology/P18-1026} {Graph-to-sequence
  learning using gated graph neural networks}.
\newblock In \emph{Proceedings of the 56th Annual Meeting of the Association
  for Computational Linguistics (Volume 1: Long Papers)}, pages 273--283.

\bibitem[{Belz et~al.(2011)Belz, White, Espinosa, Kow, Hogan, and
  Stent}]{belz2011first}
Anja Belz, Michael White, Dominic Espinosa, Eric Kow, Deirdre Hogan, and Amanda
  Stent. 2011.
\newblock The first surface realisation shared task: Overview and evaluation
  results.
\newblock In \emph{Proceedings of the 13th European workshop on natural
  language generation}, pages 217--226.

\bibitem[{Bohnet et~al.(2011)Bohnet, Mille, Favre, and
  Wanner}]{DBLP:conf/enlg/BohnetMFW11}
Bernd Bohnet, Simon Mille, Beno{\^{\i}}t Favre, and Leo Wanner. 2011.
\newblock \href {http://aclweb.org/anthology/W/W11/W11-2835.pdf} {Stumaba :
  From deep representation to surface}.
\newblock In \emph{{ENLG} 2011 - Proceedings of the 13th European Workshop on
  Natural Language Generation}, pages 232--235.

\bibitem[{Denkowski and Lavie(2014)}]{denkowski2014meteor}
Michael Denkowski and Alon Lavie. 2014.
\newblock Meteor universal: Language specific translation evaluation for any
  target language.
\newblock In \emph{Proceedings of the ninth workshop on statistical machine
  translation}, pages 376--380.

\bibitem[{Ferreira et~al.(2017)Ferreira, Calixto, Wubben, and
  Krahmer}]{ferreira2017linguistic}
Thiago~Castro Ferreira, Iacer Calixto, Sander Wubben, and Emiel Krahmer. 2017.
\newblock Linguistic realisation as machine translation: Comparing different mt
  models for amr-to-text generation.
\newblock In \emph{Proceedings of the 10th International Conference on Natural
  Language Generation}, pages 1--10.

\bibitem[{Gardent et~al.(2017{\natexlab{a}})Gardent, Shimorina, Narayan, and
  Perez-Beltrachini}]{gardent-EtAl:2017:Long}
Claire Gardent, Anastasia Shimorina, Shashi Narayan, and Laura
  Perez-Beltrachini. 2017{\natexlab{a}}.
\newblock Creating training corpora for nlg micro-planners.
\newblock In \emph{Proceedings of the 55th Annual Meeting of the Association
  for Computational Linguistics (Volume 1: Long Papers)}, pages 179--188.
\newblock (ACL 2017).

\bibitem[{Gardent et~al.(2017{\natexlab{b}})Gardent, Shimorina, Narayan, and
  Perez-Beltrachini}]{gardent-EtAl:2017:INLG2017}
Claire Gardent, Anastasia Shimorina, Shashi Narayan, and Laura
  Perez-Beltrachini. 2017{\natexlab{b}}.
\newblock The {WebNLG} challenge: Generating text from rdf data.
\newblock In \emph{Proceedings of the 10th International Conference on Natural
  Language Generation}, pages 124--133.
\newblock (INLG 2017).

\bibitem[{He et~al.(2016)He, Zhang, Ren, and Sun}]{he2016deep}
Kaiming He, Xiangyu Zhang, Shaoqing Ren, and Jian Sun. 2016.
\newblock \href {https://doi.org/10.1109/CVPR.2016.90} {Deep residual learning
  for image recognition}.
\newblock In \emph{Proceedings of the {IEEE} Conference on Computer Vision and
  Pattern Recognition, {CVPR}}, pages 770--778.

\bibitem[{Hochreiter and Schmidhuber(1997)}]{lstm1997}
Sepp Hochreiter and J{\"u}rgen Schmidhuber. 1997.
\newblock \href {https://doi.org/10.1162/neco.1997.9.8.1735} {{Long Short-Term
  Memory}}.
\newblock \emph{Neural Computation}, 9(8):1735--1780.

\bibitem[{Huang et~al.(2017)Huang, Liu, van~der Maaten, and
  Weinberger}]{DBLP:conf/cvpr/HuangLMW17}
Gao Huang, Zhuang Liu, Laurens van~der Maaten, and Kilian~Q. Weinberger. 2017.
\newblock \href {https://doi.org/10.1109/CVPR.2017.243} {Densely connected
  convolutional networks}.
\newblock In \emph{2017 {IEEE} Conference on Computer Vision and Pattern
  Recognition, {CVPR} 2017}, pages 2261--2269.

\bibitem[{Kingma and Ba(2015)}]{kingma2015adam}
Diederik~P. Kingma and Jimmy Ba. 2015.
\newblock \href {http://arxiv.org/abs/1412.6980} {Adam: {A} method for
  stochastic optimization}.
\newblock In \emph{Proceedings of the International Conference on Learning
  Representations, ICLR}.

\bibitem[{Kipf and Welling(2016)}]{kipf2016semi}
Thomas~N. Kipf and Max Welling. 2016.
\newblock \href {http://arxiv.org/abs/1609.02907} {Semi-supervised
  classification with graph convolutional networks}.
\newblock In \emph{{Proceedings of the International Conference on Learning
  Representations, ICLR}}.

\bibitem[{Klein et~al.(2017)Klein, Kim, Deng, Senellart, and Rush}]{opennmt}
Guillaume Klein, Yoon Kim, Yuntian Deng, Jean Senellart, and Alexander~M. Rush.
  2017.
\newblock \href {https://doi.org/10.18653/v1/P17-4012} {Opennmt: Open-source
  toolkit for neural machine translation}.
\newblock In \emph{Proceedings of the 55th Annual Meeting of the Association
  for Computational Linguistics, {ACL} 2017}, pages 67--72.

\bibitem[{Konstas et~al.(2017)Konstas, Iyer, Yatskar, Choi, and
  Zettlemoyer}]{konstas2017neural}
Ioannis Konstas, Srinivasan Iyer, Mark Yatskar, Yejin Choi, and Luke
  Zettlemoyer. 2017.
\newblock \href {https://doi.org/10.18653/v1/P17-1014} {Neural amr:
  Sequence-to-sequence models for parsing and generation}.
\newblock In \emph{Proceedings of the 55th Annual Meeting of the Association
  for Computational Linguistics (Volume 1: Long Papers)}, pages 146--157.

\bibitem[{Lebret et~al.(2016)Lebret, Grangier, and
  Auli}]{lebret-grangier-auli:2016:EMNLP2016}
R\'{e}mi Lebret, David Grangier, and Michael Auli. 2016.
\newblock Neural text generation from structured data with application to the
  biography domain.
\newblock In \emph{Proceedings of the 2016 Conference on Empirical Methods in
  Natural Language Processing}, pages 1203--1213.

\bibitem[{LeCun et~al.(2001)LeCun, Bottou, Bengio, and Haffner}]{lecun-01a}
Yann LeCun, Leon Bottou, Yoshua Bengio, and Patrick Haffner. 2001.
\newblock Gradient-based learning applied to document recognition.
\newblock In \emph{Proceedings of Intelligent Signal Processing}.

\bibitem[{Luong et~al.(2015)Luong, Pham, and Manning}]{luong2015effective}
Thang Luong, Hieu Pham, and Christopher~D. Manning. 2015.
\newblock \href {https://doi.org/10.18653/v1/D15-1166} {Effective approaches to
  attention-based neural machine translation}.
\newblock In \emph{Proceedings of the 2015 Conference on Empirical Methods in
  Natural Language Processing}, pages 1412--1421.

\bibitem[{Manning et~al.(2014)Manning, Surdeanu, Bauer, Finkel, Bethard, and
  McClosky}]{manning-EtAl:2014:P14-5}
Christopher Manning, Mihai Surdeanu, John Bauer, Jenny Finkel, Steven Bethard,
  and David McClosky. 2014.
\newblock The {Stanford CoreNLP} natural language processing toolkit.
\newblock In \emph{Proceedings of the 52nd Annual Meeting of the Association
  for Computational Linguistics: System Demonstrations}, pages 55--60.

\bibitem[{Marcheggiani et~al.(2018)Marcheggiani, Bastings, and
  Titov}]{marcheggiani2018exploiting}
Diego Marcheggiani, Joost Bastings, and Ivan Titov. 2018.
\newblock \href {http://arxiv.org/abs/1804.08313} {Exploiting semantics in
  neural machine translation with graph convolutional networks}.
\newblock In \emph{Proceedings of NAACL-HLT}.

\bibitem[{Marcheggiani and Titov(2017)}]{marcheggiani2017encoding}
Diego Marcheggiani and Ivan Titov. 2017.
\newblock \href {https://aclanthology.info/papers/D17-1159/d17-1159} {Encoding
  sentences with graph convolutional networks for semantic role labeling}.
\newblock In \emph{Proceedings of the 2017 Conference on Empirical Methods in
  Natural Language Processing, {EMNLP}}, pages 1506--1515.

\bibitem[{Mei et~al.(2016)Mei, Bansal, and Walter}]{mei2015talk}
Hongyuan Mei, Mohit Bansal, and Matthew~R. Walter. 2016.
\newblock \href {https://doi.org/10.18653/v1/N16-1086} {What to talk about and
  how? selective generation using lstms with coarse-to-fine alignment}.
\newblock In \emph{Proceedings of the 2016 Conference of the North American
  Chapter of the Association for Computational Linguistics: Human Language
  Technologies}, pages 720--730.

\bibitem[{Meyers et~al.(2004)Meyers, Reeves, Macleod, Szekely, Zielinska,
  Young, and Grishman}]{meyers2004nombank}
Adam Meyers, Ruth Reeves, Catherine Macleod, Rachel Szekely, Veronika
  Zielinska, Brian Young, and Ralph Grishman. 2004.
\newblock \href {http://www.lrec-conf.org/proceedings/lrec2004/pdf/398.pdf}
  {Annotating noun argument structure for nombank}.
\newblock In \emph{Proceedings of the Fourth International Conference on
  Language Resources and Evaluation, {LREC} 2004}.

\bibitem[{Mille et~al.(2017)Mille, Bohnet, Wanner, and Belz}]{mille2017shared}
Simon Mille, Bernd Bohnet, Leo Wanner, and Anja Belz. 2017.
\newblock Shared task proposal: Multilingual surface realization using
  universal dependency trees.
\newblock In \emph{Proceedings of the 10th International Conference on Natural
  Language Generation}, pages 120--123.

\bibitem[{Palmer et~al.(2005)Palmer, Gildea, and
  Kingsbury}]{palmer2005proposition}
Martha Palmer, Daniel Gildea, and Paul Kingsbury. 2005.
\newblock The proposition bank: An annotated corpus of semantic roles.
\newblock \emph{Computational linguistics}, 31(1):71--106.

\bibitem[{Papineni et~al.(2002)Papineni, Roukos, Ward, and
  Zhu}]{papineni2002bleu}
Kishore Papineni, Salim Roukos, Todd Ward, and Wei-Jing Zhu. 2002.
\newblock Bleu: a method for automatic evaluation of machine translation.
\newblock In \emph{Proceedings of 40th Annual Meeting of the Association for
  Computational Linguistics}, pages 311--318.

\bibitem[{Pennington et~al.(2014)Pennington, Socher, and
  Manning}]{pennington2014glove}
Jeffrey Pennington, Richard Socher, and Christopher Manning. 2014.
\newblock Glove: Global vectors for word representation.
\newblock In \emph{Proceedings of the 2014 conference on empirical methods in
  natural language processing (EMNLP)}, pages 1532--1543.

\bibitem[{Perez-Beltrachini et~al.(2016)Perez-Beltrachini, SAYED, and
  Gardent}]{perezbeltrachini-sayed-gardent:2016:COLING}
Laura Perez-Beltrachini, Rania SAYED, and Claire Gardent. 2016.
\newblock {Building RDF Content for Data-to-Text Generation}.
\newblock In \emph{Proceedings of COLING 2016, the 26th International
  Conference on Computational Linguistics: Technical Papers}, pages 1493--1502.

\bibitem[{Scarselli et~al.(2009)Scarselli, Gori, Tsoi, Hagenbuchner, and
  Monfardini}]{scarselli2009graph}
Franco Scarselli, Marco Gori, Ah~Chung Tsoi, Markus Hagenbuchner, and Gabriele
  Monfardini. 2009.
\newblock \href {https://doi.org/10.1109/TNN.2008.2005605} {The graph neural
  network model}.
\newblock \emph{{IEEE} Trans. Neural Networks}, 20(1):61--80.

\bibitem[{See et~al.(2017)See, Liu, and Manning}]{P17-1099}
Abigail See, Peter~J. Liu, and Christopher~D. Manning. 2017.
\newblock \href {https://doi.org/10.18653/v1/P17-1099} {Get to the point:
  Summarization with pointer-generator networks}.
\newblock In \emph{Proceedings of the 55th Annual Meeting of the Association
  for Computational Linguistics (Volume 1: Long Papers)}, pages 1073--1083.

\bibitem[{Snover et~al.(2006)Snover, Dorr, Schwartz, Micciulla, and
  Makhoul}]{snover2006study}
Matthew Snover, Bonnie Dorr, Richard Schwartz, Linnea Micciulla, and John
  Makhoul. 2006.
\newblock A study of translation edit rate with targeted human annotation.
\newblock In \emph{Proceedings of association for machine translation in the
  Americas}, volume 200.

\bibitem[{Srivastava et~al.(2014)Srivastava, Hinton, Krizhevsky, Sutskever, and
  Salakhutdinov}]{SrivastavaHKSS14}
Nitish Srivastava, Geoffrey~E. Hinton, Alex Krizhevsky, Ilya Sutskever, and
  Ruslan Salakhutdinov. 2014.
\newblock \href {http://dl.acm.org/citation.cfm?id=2670313} {Dropout: a simple
  way to prevent neural networks from overfitting}.
\newblock \emph{Journal of Machine Learning Research}, 15(1):1929--1958.

\bibitem[{Sutskever et~al.(2014)Sutskever, Vinyals, and
  Le}]{sutskever2014sequence}
Ilya Sutskever, Oriol Vinyals, and Quoc~V Le. 2014.
\newblock Sequence to sequence learning with neural networks.
\newblock In \emph{Advances in neural information processing systems}, pages
  3104--3112.

\bibitem[{Van~Noord et~al.(2018)Van~Noord, Abzianidze, Haagsma, and
  Bos}]{Noord2018LREC}
Rik Van~Noord, Lasha Abzianidze, Hessel Haagsma, and Johan Bos. 2018.
\newblock Evaluating scoped meaning representations.
\newblock In \emph{Proceedings of the Eleventh International Conference on
  Language Resources and Evaluation (LREC 2018)}.

\bibitem[{Wiseman et~al.(2017)Wiseman, Shieber, and
  Rush}]{wiseman-shieber-rush:2017:EMNLP2017}
Sam Wiseman, Stuart Shieber, and Alexander Rush. 2017.
\newblock Challenges in data-to-document generation.
\newblock In \emph{Proceedings of the 2017 Conference on Empirical Methods in
  Natural Language Processing}, pages 2243--2253.

\bibitem[{Zhang et~al.(2017)Zhang, Shrivastava, and
  Puduppully}]{DBLP:conf/eacl/ZhangSP17}
Yue Zhang, Manish Shrivastava, and Ratish Puduppully. 2017.
\newblock \href {https://aclanthology.info/papers/E17-1061/e17-1061}
  {Transition-based deep input linearization}.
\newblock In \emph{Proceedings of the 15th Conference of the European Chapter
  of the Association for Computational Linguistics, {EACL} 2017, Volume 1: Long
  Papers}, pages 643--654.

\end{thebibliography}
\bibliographystyle{acl_natbib}

\newpage

\appendix

\begin{table*}[t]
\begin{small}
 \begin{tabular}{|@{~}p{1.0cm}p{14.5cm}@{~}|}
  \hline
  WebNLG &  (Acharya Institute of Technology sportsOffered Tennis) (Acharya Institute of Technology established 2000) (Tennis sportsGoverningBody International Tennis Federation)\\  
  LSTM & The Acharya Institute of Technology was established in 2000 and is governed by the International Tennis Federation .\\
  GCN &  The sport of tennis , governed by the International Tennis Federation , is offered at the Acharya Institute of Technology which was established in 2000 .\\
  GCN$_{EC}$ & the acharya institute of technology was established in 2000 and is governed by the international tennis federation .\\  
  \hline
  WebNLG & (Acharya Institute of Technology officialSchoolColour Blue , White and Orange) (Acharya Institute of Technology was given the ' Technical Campus ' status by All India Council for Technical Education)\\
  LSTM & The Archarya Institute of Technology are blue , white and was given the Acharya Institute of Technology .\\
  GCN & The Acharya Institute of Technology was given the ' Technical Campus ' status by the All India Council for Technical Education in LOCATION . The Institute was given the " Technical Campus " status by the Acharya Institute of Technology .\\
  GCN$_{EC}$ & acharya institute of technology was given the ' technical campus ' status by the all india council for technical education which has blue , white and orange .\\
  \hline
  WebNLG &  (Saranac Lake , New York isPartOf Harrietstown , New York) (Saranac Lake , New York isPartOf Essex County , New York) (Adirondack Regional Airport cityServed Lake Placid , New York) (Adirondack Regional Airport cityServed Saranac Lake , New York)  (Saranac Lake , New York country United States)\\  
  LSTM & Adirondack Regional Airport serves the cities of Lake Placid and Saranac Lake ( Harrietstown ) in the United States . \\ 
  GCN & Adirondack Regional Airport serves the city of Saranac Lake , which is part of Harrietstown , Essex County , New York , United States . \\  
  GCN$_{EC}$ & adirondack regional airport serves the cities of lake placid and saranac lake , essex county , new york , united states . adirondack regional airport serves the city of saranac lake , essex county , new york , united states .\\  
  \hline
   WebNLG &  (Adisham Hall location Sri Lanka) (Adisham Hall architecturalStyle Tudor Revival architecture) (Adisham Hall completionDate 1931) (Adisham Hall buildingStartDate 1927)\\
   LSTM & Adisham Hall was built in 1927 and completed in 1931 . It was built in the Tudor Revival architecture style and is located in Sri Lanka .\\
   GCN & Construction of Adisham Hall , Sri Lanka began in 1927 and was completed in 1931 .\\
   GCN$_{EC}$ & adisham hall , sri lanka , constructed in 1931 , is located in sri lanka . the hall has the architectural style ' tudor revival ' .\\
  \hline
  \hline
  SR11Deep & (SROOT SROOT say) (say A0 economist) (say A1 be) (be SBJ export) (be VC think) (think A1 export) (think C-A1 have) (have VC rise) (rise A1 export) (rise A2 strongly) (strongly COORD but) (but CONJ not) (not AINV enough) (not AINV offset) (offset A1 jump) (jump A1 in) (in A1 import) (jump AINV the) (offset A2 export) (not AINV probably) (strongly TMP in) (in A1 august) (say P .)\\
  Gold & Exports are thought to have risen strongly in August , but probably not enough to offset the jump in imports , economists said . \\
  LSTM & exports said exports are thought to have rising strongly , but not enough to offset exports in the imports in august .\\
  GCN & exports was thought to have risen strongly in august but not probably to offset the jump in imports , economists said .\\
  \hline
  SR11Deep & (SROOT SROOT be) (be P ?) (be SBJ we) (be TMP be) (be SBJ project) (project A1 research) (be VC curtail) (curtail A1 project) (curtail AM-CAU to) (to A1 cut) (cut A0 government) (cut A1 funding) (funding A0 government) (to DEP due) (to R-AM-TMP when) (be VC catch) (catch A1 we) (catch A2 with) (with SUB down) (down SBJ grant) (grant AINV our) (catch P '') (catch P ``)\\
  Gold & When research projects are curtailed due to government funding cuts , are we `` caught with our grants down '' ? \\
  LSTM & is when research projects is supposed to cut `` due '' projects is caught with the grant down .\\
  GCN & when research projects are curtailed to government funding cuts due to government funding cuts , were we caught `` caught '' with our grant down ?\\  
\hline  
 \end{tabular} 
\end{small}
\vspace*{-1.5ex}
\caption{\label{tab:sys-output-additional} Examples of system output.}
\end{table*} 
\section{Supplemental Material}
\label{sec:supMaterial}

\subsection{Training details}
We implemented all our models using OpenNMT-py \cite{opennmt}.
For all experiments we used a batch size of 64 and Adam \cite{kingma2015adam} as the optimizer with an initial learning rate of 0.001.
For GCN models and baselines we used a one-layer LSTM decoder, we used dropout \cite{SrivastavaHKSS14} in  both encoder and decoder with a rate of 0.3. 
We adopt early stopping on the development set using BLEU scores and we trained for a maximum of 30 epochs.
\subsection{More example outputs}
Table \ref{tab:sys-output-additional} shows additional examples
of generated texts for source WebNLG and SR11Deep graphs.

\end{document}